\documentclass[11pt,a4paper]{article}
\usepackage[hyperref]{emnlp2020}
\usepackage{times}
\usepackage{latexsym}
\usepackage{amsmath}
\usepackage{amssymb}
\usepackage{tabularx}
\usepackage{placeins} 
\usepackage{graphicx}

\usepackage{tikz}

\usetikzlibrary{fit,positioning, matrix,decorations.pathreplacing,calc,fit,backgrounds}

\definecolor{coolgrey}{rgb}{0.55, 0.57, 0.67}

\usepackage{microtype}

\aclfinalcopy 

\title{Modulated Fusion using Transformer for Linguistic-Acoustic Emotion Recognition}

\author{Jean-Benoit Delbrouck \\
  
  Stanford University \\
    jeanbenoit.delbrouck@stanford.edu
  \\\And
  No\'e Tits \and St\'ephane Dupont \\
         Information, Signal and Artificial Intelligence\\
          University of Mons, Belgium\\
          \{noe.tits,  stephane.dupont\}@umons.ac.be
  \\}

\date{}

\begin{document}
\maketitle
\begin{abstract}
This paper aims to bring a new lightweight yet powerful solution for the task of Emotion Recognition and Sentiment Analysis. Our motivation is to propose two architectures based on Transformers and modulation that combine the linguistic and acoustic inputs from a wide range of datasets to challenge, and sometimes surpass, the state-of-the-art in the field. To demonstrate the efficiency of our models, we carefully evaluate their performances on the IEMOCAP, MOSI, MOSEI and MELD dataset. The experiments can be directly replicated and the code is fully open for future researches\footnote{\url{https://github.com/jbdel/modulated_fusion_transformer}}.

\end{abstract}

\section{Introduction}
Understanding expressed sentiment and emotions are two crucial factors in human multimodal language yet predicting affective states from multimedia remains a challenging task. The emotion recognition task has existed working on different types of signals, typically audio, video and text. Deep Learning techniques allow the development of novel paradigms to use these different signals in one model to leverage joint information extraction from different sources. These models usually require a fusion between modality, a crucial step to compute expressive multimodal features used by a classifier to output probabilities over the possible answers.

In this paper, we propose an architecture based on two stages: an independent sequential stage based on LSTM \cite{hochreiter1997long} where modality features are computed separately, and a second hierarchical stage based on Transformer \cite{vaswani2017attention} where we iteratively compute and fuse new multimodal representations. This paper proposes the fusion between the acoustic and linguistic features through attention modulation \cite{yu2019deep} and linear modulation \cite{dumoulin2018feature-wise}, a powerful tool to shift and scale the feature maps of one modality given the representation of another. 

The association of this horizontal-vertical encoding and modulated fusion shows really strong results across a wide range of datasets for emotion recognition and sentiment analysis. In addition to the interesting performances it offers, the modulation requires no or very few learning parameters, making it fast and easy to train. The paper is structured as follows: we first present the different researches used for comparison in our experiments in section \ref{sec:related}, we then briefly present the different datasets in section \ref{sec:datasets}. Then we carefully describe our sequential feature extraction based on LSTM in section \ref{sec:feat_ex} and the two hierarchical modulated fusion model, the Modulated Attention Transformer (MAT) and Modulated Normalization Transformer (MNT), in section \ref{sec:models}. Finally, we explain the experimental settings in section \ref{sec:settings} and report the results of our model variants in section \ref{sec:results}.

\section{Related Work}
\label{sec:related}
The presented related work is used for comparison for our experiments. We proceed to briefly describe their proposed models.

First, \citet{bagher-zadeh18} proposed a novel multimodal fusion technique called the Dynamic Fusion Graph (DFG) to study the nature of cross-modal dynamics in multimodal language. DFG contains built-in efficacies that are directly related to how modalities interact. 
 
To capture the context of the conversation through all modalities, the current speaker and listener(s) in the conversation, and the relevance and relationship between the available modalities through an adequate fusion mechanism, \citet{shenoy2020multilogue} proposed a recurrent neural network architecture that attempts to take into account all the mentioned drawbacks, and keeps track of the context of the conversation, interlocutor states, and the emotions conveyed by the speakers in the conversation.

\citet{pham2019found} presented a model that learns robust joint representations by cyclic translations between modalities (MCTN), that achieved strong results on various word-aligned human multimodal language tasks.

\citet{wang2019words} proposed the Recurrent Attended Variation Embedding Network (RAVEN) to model expressive nonverbal representations by analyzing the fine-grained visual and acoustic patterns that occur during word segments. In addition, they seek to capture the dynamic nature of nonverbal intents by shifting word representations based on the accompanying nonverbal behaviors.

But the related work that is probably the closest to ours is the Multimodal Transformer \cite{tsai2019MULT, delbrouck-etal-2020-transformer} because they also use Transformer based solutions to encode their modalities. Nonetheless, we differ in many ways. First, their best solutions and scores reported are using visual support. Secondly, they use Transformer for cross-modality encoding for every modality pairs; this equals to 6 Transformer modules (2 pairs per modality) while we only use two Transformer (one per modality). Finally, each output pairs is concatenated to go though a second stage of Transformer encoding. We also differ on how the features are extracted: they base their solution on CNN while we use LSTM. In this paper, it is important to note that we compare our results to their word-unaligned scores, as we do not use word-alignment either.

\section{Datasets}
\label{sec:datasets}

\subsection{IEMOCAP dataset}

IEMOCAP~\cite{busso2008iemocap} is a multimodal dataset of dyadic conversations of actors. The modalities recorded are Audio, Video and Motion Capture data. All conversations were segmented, transcribed and annotated with two different emotional types of labels: emotion categories (6 basic emotions~\cite{ekman1999basic} -- happiness, sadness, anger, surprise, fear, disgust -- plus frustrated, excited and neutral) and continuous emotional dimensions (valence, arousal and dominance). 

For categorical labels, the annotators could also select "other" if they found the emotion could not be described with one of the adjectives.
The categorical labels were given by 3-4 evaluators. Majority vote was used to have the final label. In case of ex aequo, it was considered not consistent in terms of inter-evaluator agreement; 7532 segments out of the 10039 segments reached agreement. 

To be comparable to previous research, we use the four categories: neutral, sad, happy, angry. Happy category is obtained by merging excited and happy labeled \cite{yoon18}, we obtain a total of 5531 utterances: 1636 happy, 1084 sad, 1103 angry, 1708 neutral. The train-test split is made according to \citet{poria2017context} as it seems to be the norm for recent works. 


\subsection{CMU-MOSI dataset}

CMU-MOSI~\cite{zadeh2016mosi} dataset is a collection of video clips containing opinions. The collected videos come from YouTube and were selected with metada using the \#vlog hashtag for \textit{video-blog} which desribes a specific type of video that often contains people expressing their opinion. The resulting dataset included clips with speakers with different ethnicities but all speaking in english. The speech was manually transcribed. These transcriptions were aligned with audio at word level. The videos were annotated in sentiment with a 7-point Likert scale (from -3 to 3) by five workers for each video using Amazon's Mechanical Turk.

\subsection{CMU-MOSEI dataset}

MOSEI~\cite{bagher-zadeh-etal-2018-multimodal} is the next generation of MOSI dataset. They also took advantage of online videos containing expressed opinions. They analyzed videos with a face detection algorithm and selected videos with only one speaker with an attention directed to the camera.

They used a set of 250 different keywords to scrape the videos and kept a maximum of 10 videos for each one with manual transcription included. The dataset was then manually curated to keep only data with good quality. It is annotated with a 7-point Likert scale as well as the six basic emotion categories~\cite{ekman1999basic}.

\subsection{MELD dataset} 

The Multimodal EmotionLines Dataset (MELD) \cite{poria2019meld} contains dialogue instances that encompasses audio and visual modality along with text. MELD has more than 1400 dialogues and 13000 utterances from Friends TV series. Multiple speakers participated in the dialogues. Each utterance in a dialogue has been labeled by any of these seven emotions: Anger, Disgust, Sadness, Joy, Neutral, Surprise and Fear. MELD also has sentiment (positive, negative and neutral) annotation for each utterance.

\section{Feature extractions}
\label{sec:feat_ex}
This sections aims to describe the linguistic and acoustic features used as the input of our proposed modulated fusions based on Transformers. The extraction is performed independently for each sample of a dataset. We denote the extracted linguistic features as $x$ and acoustic as $y$. In the end, both $x$ and $y$ have a size $[T, C]$ where $T$ is the temporal axis size and $C$ the feature size. Its important to note that $T$ is different for each sample, while $C$ is a hyper-parameter. 

\subsection{Linguistic}
A sentence is tokenized and lowercased. We remove special characters and punctuation. We build our vocabulary against the train-set of the datasets and embed each word in a vector of 300 dimensions using GloVe \cite{pennington2014glove}. If a word from the validation or test-set is not in present our vocabulary, we replace it with the unknown token "unk". Each sentence is run through an unidirectional one-layered LSTM of size $C$. The size of each linguistic example $x$ is therefore $[T, C]$ where $T$ is the number of words in the sentence.

\subsection{Acoustic features}

In the litterature of multimodal emotion recognition, many works use hand designed acoustic features sets that capture information about prosody and vocal quality such as ComPaRe (Computational Paralinguitic Challenge) feature sets from Interspeech conference.

However, with the evolution of deep learning models, lower level features such as mel-spectrograms have shown to be very powerful for speech related tasks such as speech recognition and speech synthesis. In this work we extract mel-spetrograms with the same procedure as a typical seq2seq Text-to-Speech system.

Specifically, our mel-spectrograms were extracted with the same procedure as in~\cite{dctts-18-tachibana} with librosa python library~\cite{librosa-15-mcfee} with 80 filter banks (the embedding size is therefore 80). A temporal reduction is then applied by selecting one frame every 16 frames. Each spectrogram is then run through an unidirectional one-layered LSTM of size $C$. The size of each acoustic example $y$ is therefore $[T, C]$ where $T$ is the number of frames in the spectrogram.

\section{Models}
\label{sec:models}

This section aims to describe the three model variants evaluated in our experiments. First, we describe the projection (P) of the features extracted in section \ref{sec:feat_ex} over emotion and sentiment classes without using any Transformer. This corresponds to the baseline for our experiments. Secondly, we present the Naive Transformer (NT) model, a transformer-based encoding where the inputs are encoded separately, the linguistic and acoustic features do not interact with each other: there is no modulated fusion. Finally, we present the two highlights of the paper, the Modulated Attention Transformer (MAT) and the Modulated Normalization Transformer (MNT), two solutions where the encoded linguistic representation modulates the entire process of the acoustic encoding.


\subsection{Projection}
\label{sec:projection}
Given the linguistic features $x$ and acoustic features $y$ extracted at section \ref{sec:feat_ex}, we define the projection as a two-step process. First, we use an attention-reduce mechanism over each modality, and then fuse both modality vectors using a simple element-wise sum.

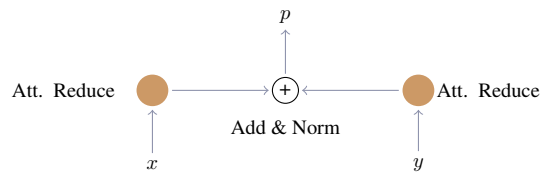
\begin{figure}[h]
\centering
\begin{tikzpicture}[scale=0.7, transform shape]
    \tikzstyle{every pin edge}=[<-,shorten <=1pt]
    \tikzstyle{neuron}=[circle,fill=brown,minimum size=17pt,inner sep=0pt]
    \tikzstyle{plus}=[circle, draw=black, inner sep=2pt,outer sep=2pt]
    \tikzstyle{func neuron}=[neuron, fill=brown!80, outer sep=2pt];

\node[func neuron] (A) at (0,0) {};
\node[left of=A, node distance=20pt,text width=10em] () {Att. Reduce};
\node[below of=A, node distance=40pt] (xin) {$x$};
\path[->,draw=coolgrey] (xin) -- (A);

\node[func neuron] (B) at (5,0) {};
\node[right of=B, node distance=65pt,text width=10em] () {Att. Reduce};
\node[below of=B, node distance=40pt] (xiny) {$y$};
\path[->,draw=coolgrey] (xiny) -- (B);  

\node[plus] (plusplus) at (2.5,0) {+};
\node[below of=plusplus, node distance=20pt] () {Add \& Norm};

\path[->,draw=coolgrey] (B) -- (plusplus);  
\path[->,draw=coolgrey] (A) -- (plusplus);  
\node[above of=plusplus, node distance=40pt] (xout) {$p$};
\path[->,draw=coolgrey] (plusplus) -- (xout);  

\end{tikzpicture}

\caption{Projection}
\label{fig:naive_fsion}
\end{figure}

The attention-reduce mechanism consists of a soft-attention over itself followed by a weighted-sum computed according to the attention weights. If we consider the feature input $x$ of size $[T, C]$:
\begin{equation}
    \begin{aligned}
        a_i &= \text{softmax}({v_i^a}^\top (W_x x)) \\
        \Bar{x} &= \sum\limits_{i=0}^{T} a_{i}x_i
    \end{aligned}
\end{equation}

After this reduce mechanism, the input becomes vectors of size $[1, C]$. We can then apply the element-wise sum as follows:

\begin{equation}
    y \sim p = W_p (\text{LayerNorm}(\Bar{x} + \Bar{y})) \label{eq:proj_layer}
\end{equation}
where $p$ is the distribution of probabilities over possible answers and LayerNorm denotes Layer Normalization \cite{ba2016layer}. If we assume the input feature $x$ has the shape $[T, C]$, for each feature channel $c \in \{1,2, \cdots, C\}$

\begin{equation}
\begin{aligned}
    \mu_{i,c} &= \frac{1}{T}  \sum_{t=1}^{T} x_{i,t,c} \\
    \sigma_{i,c}^2 &= \frac{1}{T} \sum_{t=1}^{T} (x_{i,t,c} - \mu_{i,c})^2 \\
    \hat{x}_{i,t,c} &= \frac{x_{i,t,c}-\mu_{i,c}}{\sqrt{\sigma_{i,c}^2}}
\end{aligned}
\end{equation}
Finally, for each channel, we have learnable parameters $\gamma_c$ and $\beta_c$, such that:

\begin{equation}
    y_{i,:,c} = \gamma_c \hat{x}_{i,:,c} + \beta_c \label{eq:norm_learn}
\end{equation}

\subsection{Naive Transformer}
\label{sec:mono}
The Naive Transformer model consists of stacking a Transformer on top of the linguistic and acoustic features extracted at section \ref{sec:feat_ex} before the projection of section \ref{sec:projection}. Transformers are independent and their respective input features do not interact with each other.

A Transformer is composed of a stack of $B$ identical blocks but with their own set of training parameters. Each block has two sub-layers. There is a residual connection around each of the two sub-layers, followed by layer normalization \cite{ba2016layer}. The output of each sub-layer can be written like this:
\begin{equation}
    \text{LayerNorm}(x + \text{Sublayer}(x))
\end{equation}
where Sublayer(x) is the function implemented by the sub-layer itself. In traditional Transformers, the two sub-layers are respectively a multi-head self-attention mechanism and a simple Multi-Layer Perceptron (MLP).

The attention mechanism consists of a Key $K$ and Query $Q$ that interacts together to output a attention map applied to Value $V$:
\begin{equation}
    \text{Attention}(Q, K, V) = \text{softmax}\left(\frac{QK^\top}{\sqrt{C}}\right)V
\end{equation}
In the case of self-attention, $K$, $Q$ and $V$ are the same input. If this input is of size $T \times C$, the operation $QK^\top$ results in a squared attention matrix containing the affinity between each row $T$. Expression $\sqrt{C}$ is a scaling factor. The multi-head attention (MHA) is the idea of stacking several self-attention attending the information from different representation sub-spaces at different positions:

\begin{equation}
    \begin{aligned}
    \text{MHA}(Q, K, V) = \text{Concat}(\text{head}_1, ..., \text{head}_h) W_o \\
    \text{where}\;\text{head}_i = \text{Attention}(QW_i^Q, KW_i^K, VW_i^V)
    \end{aligned}
\end{equation}

A subspace is defined as slice of the feature dimension $k$. In the case of four heads, a slice would be of size $\frac{k}{4}$. The idea is to produce different sets of attention weights for different feature sub-spaces. In the context of Transformers, $Q$, $K$ and $V$ are $x$ for the linguistic Transformer and $y$ for the acoustic Transformer. Throughout the MHA, the feature size of $x$ and $y$ remains unchanged, namely $C$. 

\begin{figure}[h]
\centering
\begin{tikzpicture}[scale=0.6, transform shape]
    \tikzstyle{every pin edge}=[<-,shorten <=1pt]
    \tikzstyle{neuron}=[circle,fill=black!25,minimum size=17pt,inner sep=0pt]
    \tikzstyle{input neuron}=[neuron, fill=green!50];
    \tikzstyle{func neuron}=[neuron, fill=purple!50];
    \tikzstyle{att neuron}=[neuron, fill=coralpink!50];
    \tikzstyle{fake neuron}=[neuron, fill=white!50];
    \tikzstyle{fusion neuron}=[neuron, fill=coolgrey!50];
    \tikzstyle{hidden neuron}=[neuron, fill=red!50];

    \tikzstyle{annot} = [text width=4em, text centered]

    \node[func neuron] (A) at (0,0) {};
    \node[annot,below of=A, node distance=70pt] (xin) {$x$};
    \path[->,draw=coolgrey] (xin) -- (A);  
    \node[fusion neuron] (B) at (0,2) {};
    \node[hidden neuron] (C) at (0,4) {};
    \node[fusion neuron] (D) at (0,6) {};

    \path[shorten >=1pt,->,draw=black!50] (A) edge (B);
    \path[shorten >=1pt,->,draw=black!50] (B) edge (C);
    \path[shorten >=1pt,->,draw=black!50] (C) edge (D);

    \draw[rounded corners] (-3, -1.5) rectangle (3.5,7) {};
    \node[annot,above right = 0.2cm and 3cm of B] () {$\times B$};

    \node[annot,below left = 1.5cm and -0.8cm of A] () {};
    \node[annot,right of=A, node distance=50pt,text width=10em] () {Multi-Head A.};
    
    \draw[draw=black!50] (0, -1) -- (-1.5, -1);
    \path[draw=black!50] (-1.5, -1) -- (-1.5, 2);
    \path[shorten >=0pt,->,draw=black!50] (-1.5, 2) -- (-0.3,2);

    \node[annot,right of=B, node distance=45pt,text width=10em] () {Add \& Norm};

    \draw[draw=black!50] (0, 3) -- (-1.5, 3);
    \path[draw=black!50] (-1.5, 3) -- (-1.5, 6);
    \path[shorten >=0pt,->,draw=black!50] (-1.5, 6) -- (-0.3,6);

    \node[annot,right of=C, node distance=25pt] () {{MLP}};
    \node[annot,right of=D, node distance=45pt,text width=10em] () {Add \& Norm};
    
    \node[annot,above of=D, node distance=50pt] (xout) {$\tilde{x}$};
    \path[->,draw=black] (D) -- (xout);

\end{tikzpicture}
\caption{Linguistic Naive Transformer.}
\label{fig:transformer}
\end{figure}
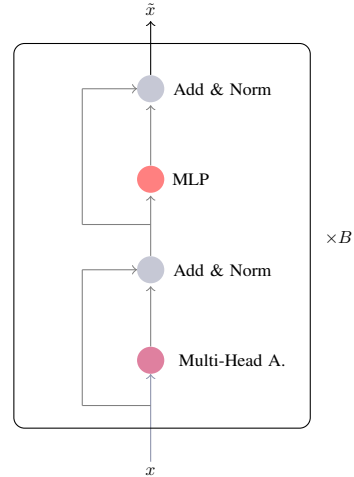

The MLP consists of two layers of respective sizes $[C \rightarrow \overline{C}]$  and $[\overline{C} \rightarrow C]$. After encoding through the blocks, the outputs $\tilde{x}$ and $\tilde{y}$ can be used by the projection layer (section \ref{sec:projection}) for classification. In Figure \ref{fig:transformer}, we show the encoding of the linguistic features $x$ and its corresponding output $\tilde{x}$.

\subsection{Modulated Fusion}

The Modulated Fusion consists of modulating the encoding of the acoustic features $y$ given the encoded linguistic features $\tilde{x}$. This modulation in the acoustic Transformer allows for an early fusion of both modality whose result is going to be $\tilde{y}$. This modulation can be performed through the Multi-Head Attention or the Layer-Normalization. After, the output $\tilde{x}$ and $\tilde{y}$ are used as input of the projection from section \ref{sec:projection}. We proceed to describe both approaches in the next sub-sections.

\subsubsection{Modulated Attention Transformer}
To modulate the acoustic self-attention by the linguistic output, we switch the key $K$ and value $V$ of the self-attention from $y$ to $\tilde{x}$. The operation $QK^\top$ results in an attention map that acts like an affinity matrix between the rows of modality matrix $\tilde{x}$ and  $y$. This computed alignment is applied over the Value $V$ (now $\tilde{x}$) and finally we add the residual connection $y$. The following equation describes the new attention sub-layer in the acoustic Transformer.

\begin{equation}
   y = \text{LayerNorm}(y + \text{MHA}(y,x,x)) \label{eq:block_output}
\end{equation}

For the operation $QK^\top$ to work as well as the residual connection (the addition), the feature sizes $C$ of $\tilde{x}$ and $y$ must be equal. This can be adjusted with the different transformation matrices of the MHA module or the LSTM size of section \ref{sec:feat_ex}.

\begin{figure}[h]
\centering
\begin{tikzpicture}[scale=0.55, transform shape]
    \tikzstyle{every pin edge}=[<-,shorten <=1pt]
    \tikzstyle{neuron}=[circle,fill=black!25,minimum size=17pt,inner sep=0pt]
    \tikzstyle{input neuron}=[neuron, fill=green!50];
    \tikzstyle{func neuron}=[neuron, fill=purple!50];
    \tikzstyle{att neuron}=[neuron, fill=coralpink!50];
    \tikzstyle{fake neuron}=[neuron, fill=white!50];
    \tikzstyle{fusion neuron}=[neuron, fill=coolgrey!50];
    \tikzstyle{hidden neuron}=[neuron, fill=red!50];

    \tikzstyle{annot} = [text width=4em, text centered]
    
    \node[annot,below of=A, node distance=60pt] (xin) {$x$};
    \path[->,draw=coolgrey] (xin) -- (A);  
    \node[func neuron] (A) at (0,0) {};
    \node[fusion neuron] (B) at (0,2) {};
    \node[hidden neuron] (C) at (0,4) {};
    \node[fusion neuron] (D) at (0,6) {};

    \path[shorten >=1pt,->,draw=black!50] (A) edge (B);
    \path[shorten >=1pt,->,draw=black!50] (B) edge (C);
    \path[shorten >=1pt,->,draw=black!50] (C) edge (D);

    \draw[rounded corners] (-2.5, -1.5) rectangle (3.5,7) {};

    \node[annot,right of=A, node distance=50pt,text width=10em] () {Multi-Head A.};
    
    \draw[draw=black!50] (0, -1) -- (-1.5, -1);
    \path[draw=black!50] (-1.5, -1) -- (-1.5, 2);
    \path[shorten >=0pt,->,draw=black!50] (-1.5, 2) -- (-0.3,2);

    \node[annot,right of=B, node distance=45pt,text width=10em] () {Add \& Norm};

    \draw[draw=black!50] (0, 3) -- (-1.5, 3);
    \path[draw=black!50] (-1.5, 3) -- (-1.5, 6);
    \path[shorten >=0pt,->,draw=black!50] (-1.5, 6) -- (-0.3,6);

    \node[annot,right of=C, node distance=25pt] () {{MLP}};
    \node[annot,right of=D, node distance=45pt,text width=10em] () {Add \& Norm};
    
    \node[annot,above of=D, node distance=50pt] (xout) {$\tilde{x}$};
    \path[->,draw=black] (D) -- (xout);


    \node[func neuron] (A) at (0+7,0) {};
    \node[annot,below of=A, node distance=60pt] (yin) {$y$};
    \path[->,draw=coolgrey] (yin) -- (A); 
    \node[fusion neuron] (B) at (0+7,2) {};
    \node[hidden neuron] (C) at (0+7,4) {};
    \node[fusion neuron] (D) at (0+7,6) {};

    \path[shorten >=1pt,->,draw=black!50] (A) edge (B);
    \path[shorten >=1pt,->,draw=black!50] (B) edge (C);
    \path[shorten >=1pt,->,draw=black!50] (C) edge (D);

    \draw[rounded corners] (-2.5+7, -1.5) rectangle (3.5+7,7) {};
    \node[annot,above right = 0.2cm and 3cm of B] () {$\times B$};

    \node[annot,right of=A, node distance=50pt,text width=10em] () {Multi-Head A.};
    
    \draw[draw=black!50] (0+7, -1) -- (-1.5+7, -1);
    \path[draw=black!50] (-1.5+7, -1) -- (-1.5+7, 2);
    \path[shorten >=0pt,->,draw=black!50] (-1.5+7, 2) -- (-0.3+7,2);

    \node[annot,right of=B, node distance=45pt,text width=10em] () {Add \& Norm};

    \draw[draw=black!50] (0+7, 3) -- (-1.5+7, 3);
    \path[draw=black!50] (-1.5+7, 3) -- (-1.5+7, 6);
    \path[shorten >=0pt,->,draw=black!50] (-1.5+7, 6) -- (-0.3+7,6);

    \node[annot,right of=C, node distance=25pt] () {{MLP}};
    \node[annot,right of=D, node distance=45pt,text width=10em] () {Add \& Norm};
    
    \node[annot,above of=D, node distance=50pt] (xout) {$\tilde{y}$};
    \path[->,draw=black] (D) -- (xout);


    \draw[draw=blue] (0.5,7.7) -- (4, 7.7);
    \draw[draw=blue] (4, 7.7) -- (4, 0);
    \draw[->,draw=blue] (4, 0) -- (6.5, 0);

\end{tikzpicture}
\caption{Modulated Attention Transformer.}
\label{fig:modulated_attention}
\end{figure}
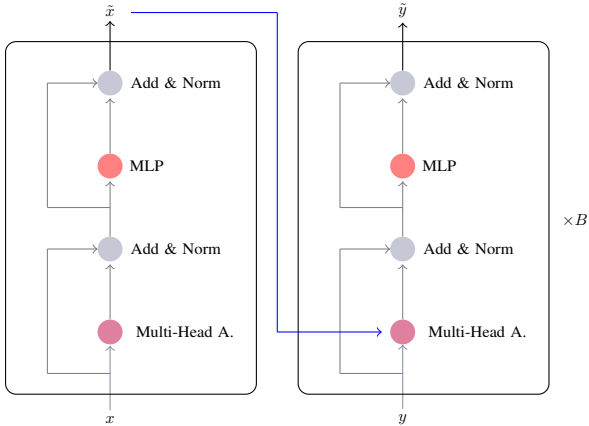

If we consider that $\tilde{x}$ is of size $[T_x, C]$ and $y$ of size $[T_y, C]$, then the sizes of the matrix multiplication operations of this modulated attention can be written as follows (where $\times$ denotes matrix multiplication):

\begin{align}
    y \times x^T &= T_y, C \times C, T_x = T_y, T_x \label{eq:att1} \\
    (\ref{eq:att1}) \times x &= T_y, T_x \times T_x, C = T_y, C \label{eq:att2} \\
    (\ref{eq:att2}) + y &= T_y, C + T_y, C = T_y, C \label{eq:att3} 
\end{align}

where equation \ref{eq:att3} denotes the $(y + \text{MHA}(y,x,x))$ operation. 

We call the Modulated Attention Transformer "MAT" in the experiments.

\subsubsection{Modulated Normalization Transformer}
It is possible to modulate the normalization layers by predicting two scalars per block from $\tilde{x}$, namely $\Delta \gamma$ and $\Delta \beta$, that will be added to the learnable parameters of equation \ref{eq:norm_learn}:

\begin{equation}
\begin{aligned}
    \overline{\gamma_c} = \gamma_c + \Delta \gamma \\
    \overline{\beta_c} = \beta_c + \Delta \beta
\end{aligned}
\end{equation}

where $\Delta \gamma,\; \Delta \beta  = \text{MLP}(\tilde{x})$ and the MLP has one layer of sizes $[C, 4 \times B]$. Two pairs of scalars per block are predicted, so no scalars are shared amongst normalization layers.

We update the layer normalization equation accordingly:
\begin{equation}
    y_{i,:,c} = \overline{\gamma_c} \hat{x}_{i,:,c} + \overline{\beta_c} 
\end{equation}

The Modulated Normalization is a computationally efficient and powerful method to modulate neural activations. It enables the linguistic output to manipulate entire acoutisc feature maps by scaling them up or down, negating them, or shutting them off. As there is only two parameters per feature map, the total number of new training parameters is small. This makes the Modulated Normalization a very scalable method.

We call the Modulated Normalization Transformer "MNT" in the experiments.

\section{Experimental settings}
\label{sec:settings}

We train our models using the Adam optimizer \cite{kingma2014adam} with a learning rate of $1e-4$ and a mini-batch size of 32. If the accuracy score on the validation set does not increase for a given epoch, we apply a learning-rate decay of factor $0.5$. We decay our learning rate up to 2 times. Afterwards, we use an early-stop of 10 epochs on accuracy. Results presented in this paper are from the averaged predictions of at most 10 models. 

Unless stated otherwise, the LSTM size $C$ (and therefore the Transformer size) is 512. We use $B=2$ Transformer blocks for P and NT models and $B=4$ for MNT and MAT models. We use 8 multi-heads regardless of the models or the modality encoded. The size $\overline{C}$ of the Transformer MLP is set at 2048. We apply dropout of 0.1 on the output of each block iteration, and 0.5 on the input $(x+y)$ of the projection layer (equation \ref{eq:proj_layer}). 

\section{Results}
\label{sec:results}

We present the results on four sentiment and emotion recognition datasets: IEMOCAP, MOSEI, MOSI and MELD. For each dataset, the results are presented in terms of the popular metrics used for the dataset. Most of the time, F1-score is used, and sometimes the weighted F1-scores to take into account the imbalance between emotion or sentiment classes.

\textbf{IEMOCAP} \quad We first compare the precision, recall and unweighted F1-scores of our two model variants on IEMOCAP in Table \ref{table:iemocap}. We notice that our MAT model comes on top.

\begin{table}[ht]
	\centering
	\resizebox{\linewidth}{!}{
	\begin{tabular}{lcccccc}
	Model  & Prec. &  Recall & F1 \\
    \hline
    MAT (L+A, ours) & \textbf{0.74} & \textbf{0.74} & \textbf{0.74} \\
    MNT (L+A, ours) & 0.72 & 0.72 & 0.72 \\
    NT (L+A, ours) & 0.71 & 0.70 & 0.70 \\
    P (L+A, ours) & 0.69 & 0.67 & 0.67 \\
    Mult (L+A+V, \citeyear{tsai2019MULT}) & - & - & 0.715 \\
    E2 (L+A, \citeyear{sahu2019multimodal}) & 0.73 & 0.715 & 0.72  \\
    MDRE (L+A, \citeyear{yoon18}) & 0.72 & - & -\\
    MDREA (L+A, \citeyear{yoon18}) & 0.69 & - & -\\
    E1 (L+A, \citeyear{sahu2019multimodal}) & 0.73 & 0.655 & 0.68  \\
    RAVEN, (L+A+V, \citeyear{wang2019words}) & - & - & 0.665  \\
    MCTN, (L, \citeyear{pham-etal-2018-seq2seq2sentiment}) & - & - & 0.66
	\end{tabular}}
\caption{Results of the 4-emotions task of IEMOCAP. Prec. stands for precision and F1 is the unweighted F1-score.}
\label{table:iemocap_prec_recall}
\end{table}

If we compare the F1-score per class (table \ref{table:iemocap_prec_recall_f1}), we notice that our model MAT outperforms previous researches, the biggest margin being in the happy category. The model MulT \cite{tsai2019MULT} still comes on top in the neutral category.

\begin{table}[h]
	\centering
	\resizebox{\linewidth}{!}{
	\begin{tabular}{lcccccc}
	Model  & Hap. & Ang. & Sad & Neu. & avg\\
    \hline
    MAT (ours) & \textbf{0.68} & \textbf{0.71} & \textbf{0.75} & 0.80 & \textbf{0.73} \\
    MNT (ours) & 0.66 & \textbf{0.71} & 0.72 & 0.80 & 0.72  \\
    NT (ours) & 0.67 & 0.69 & 0.69 & 0.78 & 0.70  \\
    P (L+A, ours) & 0.63 & 0.65 & 0.68 & 0.78 & 0.67 \\
    MulT (\citeyear{tsai2019MULT}) & 0.60 & 0.70 & 0.74 & \textbf{0.82} & 0.71 \\
    MCTN (\citeyear{pham-etal-2018-seq2seq2sentiment}) & 0.49 & 0.66 & 0.72 & 0.78 & 0.66 \\
    RAVEN (\citeyear{wang2019words}) & 0.60 & 0.64 & 0.66 & 0.77 & 0.66  
	\end{tabular}}
\caption{IEMOCAP: F1-scores per emotion class. Avg denotes the weighted average F1-score.}
\label{table:iemocap_prec_recall_f1}
\end{table}

\begin{figure}[!h]
  \centering
\includegraphics[width=\linewidth]{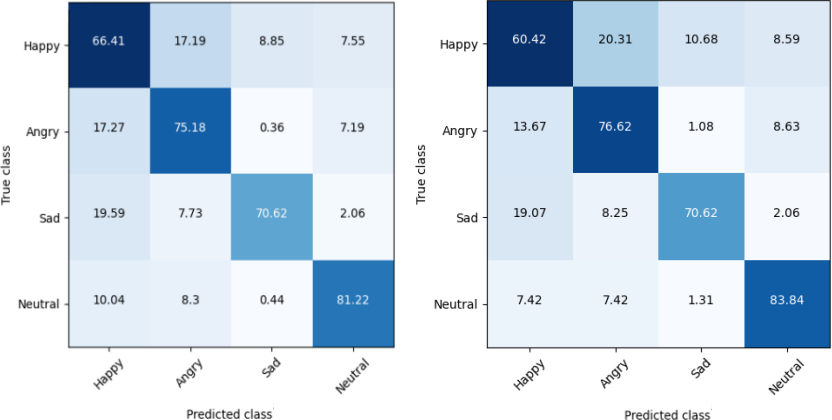}
\caption{Confusion matrices for IEMOCAP emotion task.}
\label{fig:iemocap_confusion}
\end{figure}

We can see in Figure \ref{fig:iemocap_confusion} that our MNT model has a really good recall on the neutral category but MAT significantly outperforms MNT in the happy cateogry. However, we can see that the happy class surprisingly remains a challenge for the models presented. Our MAT model predicted around 17\% of the time "angry" when the true class was happy. On the contrary, our model predicted "happy" 19\% of the time when the true label was "sad" and 17\% of the time when the true class was "angry". We can see that this is still a significant margin of error for such contradictory labels. It shows that visual cues might be necessary to further improve the performances.

\textbf{MOSI} \quad MOSI is a small dataset with few training examples. To train such models, regularization is usually needed to not overfit the training-set. In our case, dropout was enough to top the state-of-the-art results on this dataset.

Even if the dataset is a bit unbalanced between the binary answers (positive and negative), weighting the loss accordingly did not improve the results. It shows that our model variants manage to efficiently discriminate between both classes.

\begin{table}[h]
	\centering
	\begin{tabular}{lcccccc}
	Model  & F1 \\
    \hline
    MAT (L+A, ours) & {0.80} (0.84 / 0.73) \\
    MNT (L+A, ours) & {0.80} (0.84 / 0.73) \\
    NT (L+A, ours) & {0.78} (0.83 / 0.71) \\
    P (L+A, ours) & {0.76} (0.80 / 0.71) \\
    MulT (L+A+V, \citeyear{tsai2019MULT}) & \textbf{0.81} \\
    SA-Gating B6 (L+A+V, \citeyear{kumar2020gated}) & \textbf{0.81} \\
    Multilogue-Net (L+A+V, \citeyear{shenoy2020multilogue}) & 0.80 \\
    Multilogue-Net (L+A, \citeyear{shenoy2020multilogue}) & 0.79 \\
	\end{tabular}
\caption{Results on the 2-sentiment task of MOSI. Results given are the weighted F1-scores.}
\label{table:iemocap}
\end{table}

\FloatBarrier


\textbf{MOSEI} \quad MOSEI is a relatively large-scale dataset. We expect to see a more noticeable difference of score between our Modulated Transformer variants and the Naive Transformer and Projection baselines. 

\begin{table}[h]
	\centering
	\begin{tabular}{lcccccc}
	Model  & Happy & Sad & Angry \\
    \hline
    MNT (ours) & 0.66 & \textbf{0.76} & 0.77 \\
    MAT (ours) & 0.66 & 0.75 & 0.75  \\ 
    NT (ours) & {0.65} & 0.75 & 0.74  \\ 
    M-logue (\citeyear{shenoy2020multilogue}) & \textbf{0.68} &\ 0.75 & \textbf{0.81 } \\
    G-MFN (\citeyear{bagher-zadeh18}) & 0.66 & 0.67 & 0.73 \\
	Model  & Fear & Disgust & Surprise \\
    \hline 
    MNT (ours) & \textbf{0.92} & 0.85 & \textbf{0.91} \\
    MAT & 0.91 & 0.84 & 0.89\\
    P (ours) & {0.88} & 0.84 & 0.86  \\ 
    Multilogue & 0.87 & \textbf{0.87} & 0.81\\
    G-MFN & 0.79 & 0.77 & 0.85
	\end{tabular}
\caption{Results on the 6-emotions classification task of MOSEI. Metrics reported are the weighted F1-scores. M-logue stands for Multilogue-Net and G-MFN for Graph-MFN.}
\label{table:iemocap_emotion}
\end{table}

For the emotion task in Table \ref{table:iemocap_emotion}, MNT comes on top with a noticeable improvement over the state-of-the-art in the Surprise and Fear category. Multilogue still shows strong results in the Happy and Angry category, two important classes of the MOSEI dataset as they have the biggest support (respectively 2505 and 1071 samples over 6336 in the test-set). For binary sentiment classification (Table \ref{table:iemocap_sent}),  MAT is the strongest reported model.

\begin{table}[h]
	\centering
	\begin{tabular}{lcccccc}
	Model  & A2 & F1 \\
    \hline
    MAT (L+A, ours) &  \textbf{0.82} & \textbf{0.82} \\
    MNT (L+A, ours) &  {0.805} & {0.805} \\
    NT (L+A, ours) &  {0.81} & {0.80} \\
    P (L+A, ours) &  {0.805} & {0.79} \\
    MulT (L+A+V, \citeyear{tsai2019MULT}) & 0.815 & 0.815 \\
    RAVEN (L+A+V, \citeyear{wang2019words}) &  0.79 &  0.795 \\
    G-MFN (L+A+V, \citeyear{bagher-zadeh18}) & 0.79  & -\\
    MCTN (L, \citeyear{pham2019found}) & 0.75 & 0.76 
	\end{tabular}
\caption{Results on the 2-sentiments task of MOSEI. Results given are the accuracies and weighted F1-scores.}
\label{table:iemocap_sent}
\end{table}

\textbf{MELD} \quad  MELD is a dataset for Emotion Recognition in Conversation. Even if our approaches do not take into account the context, we can see that it leads to interesting results. More precisely, our variants are able to detect difficult emotion, such as fear and disgust, even though they are present in very low quantity in the training and test-set.

\begin{figure}[!h]
  \centering
\includegraphics[width=\linewidth]{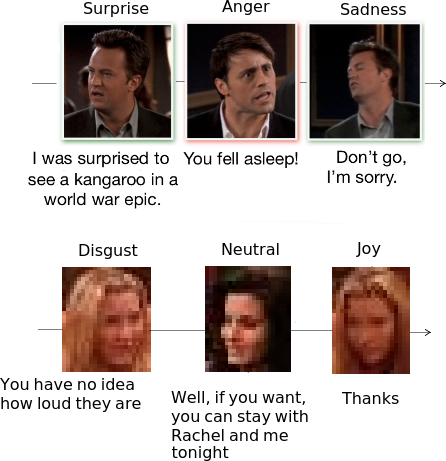}
\caption{MELD: Two contextual examples with three training samples each.}
\label{fig:meld}
\end{figure}

We can see in Table \ref{table:meld} that even if we do not use the contextual nor the speaker information, our models achieve good results in two categories: fear and disgust. To help understand these results, we give two MELD examples in Figure \ref{fig:meld}. In the top example, it is unlikely to answer "anger" to the sentence "you fell asleep!" without context, it could be surprise or fear. This is why our "anger" score is really low. In the bottom example, "you have no idea how loud they are" could very well be "anger" too, but happens to be labeled "disgust".


\begin{table}[h]
	\centering
	\begin{tabular}{lcccccc}
	Model  & Ang. & Dis. & Fear & Joy \\
    \hline
    MNT (ours) &  0.27 & \textbf{0.21} & \textbf{0.12} & 0.41 \\
    MAT (ours) & 0.27 & 0.15 & 0.09 & 0.42 \\
    NT (ours) &  0.25 & {0.11} & {0.05} & 0.39 \\
    CGCN*$\dagger$ (\citeyear{zhang2019modeling}) & \textbf{0.47} & 0.11 & 0.09 & 0.53 \\   
    DRNN* (\citeyear{majumder2019dialoguernn}) & 0.46 & 0.0 & 0.0 & 0.53 \\
    BC-LSTM* & 0.46 & 0.0 & 0.0 & 0.50  \\
    G-MFN (\citeyear{zadeh2018memory}) & 0.40 & 0.0 & 0.0 & 0.47  \\         
    Model  & Neut. & Sad & Surp. \\
    \hline
    MNT (ours) &  0.66 & 0.24 & 0.46\\
    MAT (ours) & 0.63 & 0.22 & 0.44\\
    NT (ours) &  0.54 & 0.21 & 0.41 \\
    GCN*$\dagger$ & \textbf{0.77} & \textbf{0.28 }& {0.50}  \\ 
    DRNN*  & 0.73 & 0.25 &\textbf{0.52} \\
    BC-LSTM* & 0.76 & 0.16 & 0.48 \\
    G-MFN & 0.76 & 0.13 & 0.41
	\end{tabular}
\caption{Results of the 7-emotions (Anger, Disgust, Fear, Joy, Neutral, Sad, Surprise) task of MELD. Results given in term of F1-scores. DRNN is DialogueRNN, G-MFN is Graph-MFN and CGCN is ConGCN. * denotes that a model uses the contextual information and $\dagger$ speaker information.}
\label{table:meld}
\end{table}

It is possible that our model, without any prior or contextual bias about an utterance, classify sentences similar to "you fell asleep" or "you have no idea how" as "disgust" or "fear". Further analysis on why our model perform so well could shed the light on this odd behavior. We also fall short on the sad and surprise category compared to GCN, showing that a variant of our proposed models that takes into account the context could lead to competitive results.

\section{Further analysis}
\label{sec:analysis}

A few supplementary comments can be made about the results. First, we notice that the hierarchical structure of the network brought by the transformers did bring improvements across all datasets. Indeed, even the NT model does bring significant performances boost compared to the P model that only consists of an LSTM and the projection layer. A very nice property of our solutions is that few Tranformers layers are required to be the found settings. It usually varies from 2 to 4 layers, allowing our solutions to converge very rapidly.

\begin{table}[h]
	\centering
	\begin{tabular}{lcccccc}
	   MOSEI & Params & s/epoch & epoch/c \\
	   \hline
	   P & 9.8 M & 10 & 2\\
	   NT $B=2$ & 22.9 M & 26 & 6 \\
   	   NT $B=4$ & 35.5 M & 42 & 7\\
   	   MAT $B=2$ & 22.9 M & 26 & 8\\
   	   MAT $B=4$ & 35.5 M & 42 & 10\\
   	   MNT $B=2$ & 24.5 M & 26 & 6\\
   	   MNT $B=4$ & 39.9 M & 44 & 8
   	   
	\end{tabular}
\caption{Results on a single GTX 1080 Ti for $C=512$. The statistics reported are from the MOSEI dataset for the sentiment task, as it contains the most training samples (16320). s/epoch means seconds per epoch and epoch/c means the number of epoch to convergence. Parameters are reported in Million.}
\label{table:stats}
\end{table}

Another point is that the MAT variant does not require additional training parameters nor computational power (as shown in Table \ref{table:stats}), the solution only switch one input of the Multi-Head Attention from one modality matrix to another. For MNT, the Transformer block implements only 2 normalization layers, therefore the conditional layer must only compute 2048 scalars (given $C$ is 512) for $\Delta \gamma$ and $\Delta \beta$ or roughly 1 Million parameters per block. This solution grows linearly with the hidden size but we got better results with $C = 512$ rather than 1024.

\begin{figure}[!h]
  \centering
\includegraphics[width=\linewidth]{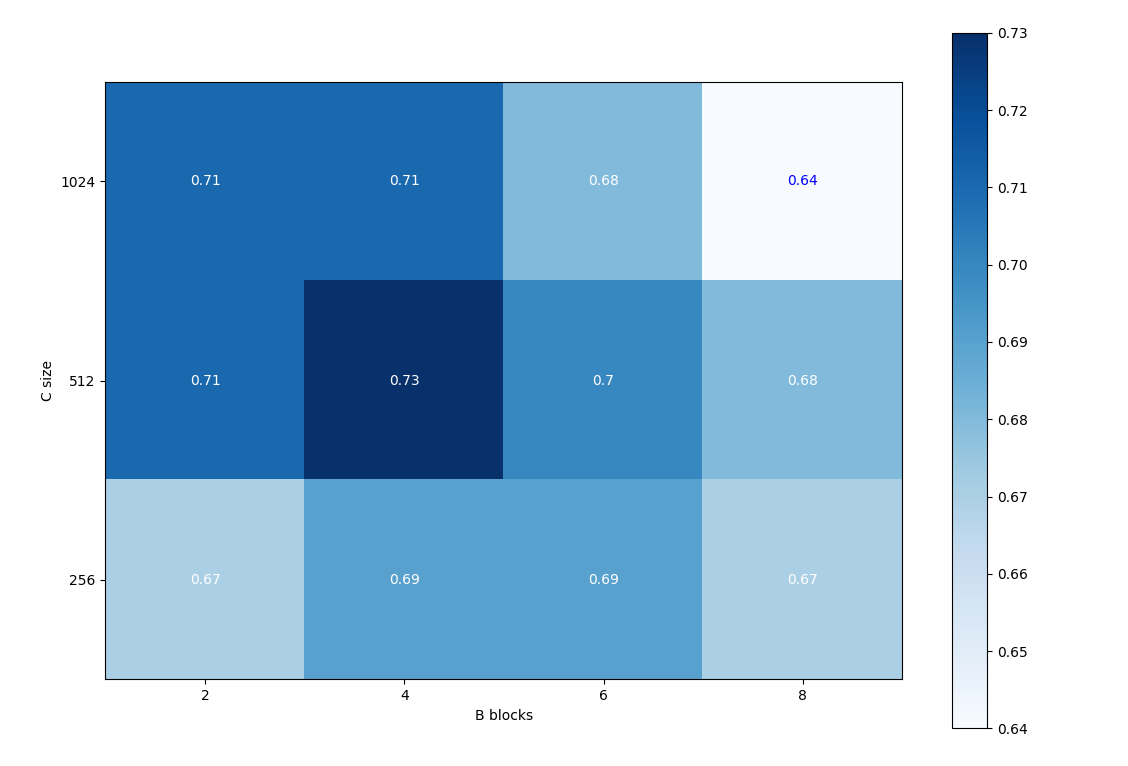}
\caption{Heatmap showing the influence on f1-scores from parameters $B$ and $C$ on IEMOCAP.}
\label{fig:heatmap}
\end{figure}

The difference between MAT and MNT variant is slim, but it seems that MAT is more suitable for the binary sentiment classification. The computed alignment by the modulated attention of the linguistic and acoustic modality proves to be an acceptable solution for 2-class problem, but seems to fall short for more nuanced classification such as multi-class emotion recognition. MNT seems more suitable for that task, as shown for MOSEI and MELD. A potential issue for MAT is that we work with shallow architectures ($B = 4$) compared to recent NLP solutions like BERT using up to 48 layers. In the scope of the dataset presented, we have not enough samples to train such architectures. It is possible that MNT adjust better with shallow layers because it can modulate entire feature maps twice per blocks. 

\section{Conclusions}

In this paper, we propose two different architectures, MAT (Modulated Attention Transformer) and MNT (Modulated Normalization Transformer), for the task of emotion recognition and sentiment analysis. They are based on Transformers and use two modalities: linguistic and acoustic. \\

The performance of our methods were thoroughly studied by comparison with a Naive Transformer baseline and the most relevant related works on several datasets suited for our experiments. 

We showed that our Transformer baseline encoding separately both modalities already performs well compared to state-of-the-art. The solutions including modulation of one modality from the other show a higher performance. Overall, the architectures offer an efficient, lightweight and scalable solution that challenges, and sometimes surpasses, the previous works in the field.

\section{Acknowledgements}

No\'e Tits  is funded through a FRIA grant (Fonds pour la Formation \`a la Recherche dans l'Industrie et l'Agriculture, Belgium). 

\bibliography{emnlp2020}
\bibliographystyle{acl_natbib}

\end{document}